\documentclass{article}

\usepackage{spconf,amsmath,amssymb,graphicx,booktabs}
\usepackage[hidelinks]{hyperref}
\usepackage{float}
\usepackage{eso-pic}

\title{DiaWhisper-DPO: Role-Attributed Transcription of Clinical Interviews via Failure-Mined Preference Optimization}

\name{
Weiming Li\textsuperscript{1},
Ana Catarina Fidalgo Barata\textsuperscript{1},
Miguel Constante\textsuperscript{2},
and Jo\~{a}o Miguel Sanches\textsuperscript{1}
}

\address{
\textsuperscript{1}Institute for Systems and Robotics (ISR), LARSyS, Departamento de Bioengenharia,\\
Instituto Superior T\'ecnico (IST), Universidade de Lisboa, 1049-001 Lisboa, Portugal\\
\textsuperscript{2}Hospital Beatriz Ângelo, Faculdade de Medicina Universidade Católica Portuguesa,
2674-514 Loures, Portugal
}

\begin{document}

\AddToShipoutPictureFG*{%
  \AtPageUpperLeft{%
    \raisebox{-8mm}[0pt][0pt]{%
      \makebox[\paperwidth][c]{%
        \parbox{0.94\paperwidth}{%
          \centering
          \fontsize{8.5}{9.5}\selectfont
          This work has been submitted to the IEEE for possible publication.
          Copyright may be transferred without notice, after which this version
          may no longer be accessible.
        }%
      }%
    }%
  }%
}

\maketitle

\begingroup
\renewcommand{\thefootnote}{}
\footnotetext{\scriptsize
This work was supported by LARSyS funding
(DOI: 10.54499/LA/P/0083/2020 and 10.54499/UID/50009/2025).}
\addtocounter{footnote}{-1}
\endgroup

\begin{abstract}
Automated depression screening from clinical interviews requires attribution of utterances to the clinician or patient. We evaluate two datasets: DAIC-WOZ, where participant-only recordings require re-synthesizing both sides for controlled two-party evaluation, and PDCH-HAMD, comprising voice-converted real Chinese interviews for cross-lingual validation. Cascaded systems combine speaker diarization with role-assignment heuristics, so errors can propagate across stages. We propose an end-to-end model, which we named \textbf{DiaWhisper}, that fine-tunes Whisper-large-v3 with LoRA and an auxiliary frame-level role head for transcription and attribution, together with \textbf{DiaWhisper-DPO}, a failure-mined refinement that uses genuine decoding failures as DPO rejected completions without human preference annotation. On 29 DAIC-WOZ test sessions, DiaWhisper-DPO achieves 0.973 role accuracy and 0.119 DER, 72\% below the strongest cascaded baseline, and reduces seed variation from $\sigma=.205$ to $.002$. Retrained on PDCH-HAMD, it achieves 0.757 role accuracy and improves all 78 session-seed pairs.
\end{abstract}


\section{Introduction}
\label{sec:intro}
\vspace{-0.25cm}

Automated depression screening from clinical interviews depends on more than accurate transcription: it also requires knowing \emph{who} said what. Patient word choice, hesitation, response latency, and turn-taking dynamics are established markers of depressive symptomatology, but they are informative only when correctly attributed and often depend on the preceding interviewer question \cite{alhanai2018detecting,gong2017topicmodeling}. The same lexical or acoustic pattern can have different clinical relevance depending on whether it comes from the clinician or patient, and misattributed turns distort downstream screening signals \cite{burdisso2024daicwoz}, making reliable role attribution clinically important for downstream clinical analysis, where speaker identity directly conditions interpretation. We study this problem on two complementary corpora: DAIC-WOZ \cite{gratch2014daic,ringeval2019avec}, a widely used English corpus whose participant-only recordings require two-party re-synthesis for controlled evaluation (Section~\ref{ssec:dataset}), and PDCH-HAMD \cite{pdch2025}, voice-converted recordings of real Chinese clinical interviews for independent cross-lingual validation (Section~\ref{ssec:setup}). Traditional pipelines use speaker diarization \cite{park2022diarization} followed by heuristic role assignment, but this cascaded design lets errors propagate: diarization mistakes carry into role assignment, while neither stage can correct the other \cite{elshafey2019jointdiarization}. Recent work instead models transcription and role or speaker attribution jointly within a single sequence-transduction model \cite{elshafey2019jointdiarization,kanda2020sot,xu2026jointasr}, alongside parameter-efficient Whisper adaptation \cite{song2024lorawhisper} and preference optimization for code-switching ASR \cite{nguyenquang2026dpocs}. We combine these directions by fine-tuning Whisper to produce role-tagged, timestamped transcripts end-to-end, then refining it with failure-mined DPO \cite{rafailov2023dpo,ouyang2022instructgpt}, using genuine decoding failures as rejected completions and ground truth as chosen completions, without additional preference annotation.

Our contributions are: (i) \textbf{failure-mined preference optimization}, showing that genuine model failures provide a stronger and more stable training signal than extra SFT or artificial negatives, without additional annotation cost; (ii) \textbf{DiaWhisper}, a LoRA-tuned Whisper-large-v3 with an auxiliary frame-level role-classification head for role-tagged, timestamped two-party transcription without separate diarization; (iii) a DAIC-WOZ real-versus-synthetic acoustic validation that quantifies the participant-side domain gap introduced by re-synthesis; (iv) \textbf{DiaWhisper-DPO}, which on DAIC-WOZ improves seed stability from $\sigma$=.205 to .002 and reduces DER by 72\% relative to the strongest cascaded baseline; when independently trained and evaluated on voice-converted PDCH-HAMD, the same method improves role attribution in all 78 matched session-seed cases, supporting its cross-lingual applicability; and (v) per-role AER and DER with bootstrap confidence intervals, coverage-conditioned accuracy, and oracle role-mapping checks, which expose role- and language-specific failures while separating genuine attribution gains from coverage or heuristic effects.

\begin{figure*}[t]
\centering
\includegraphics[width=0.8\textwidth]{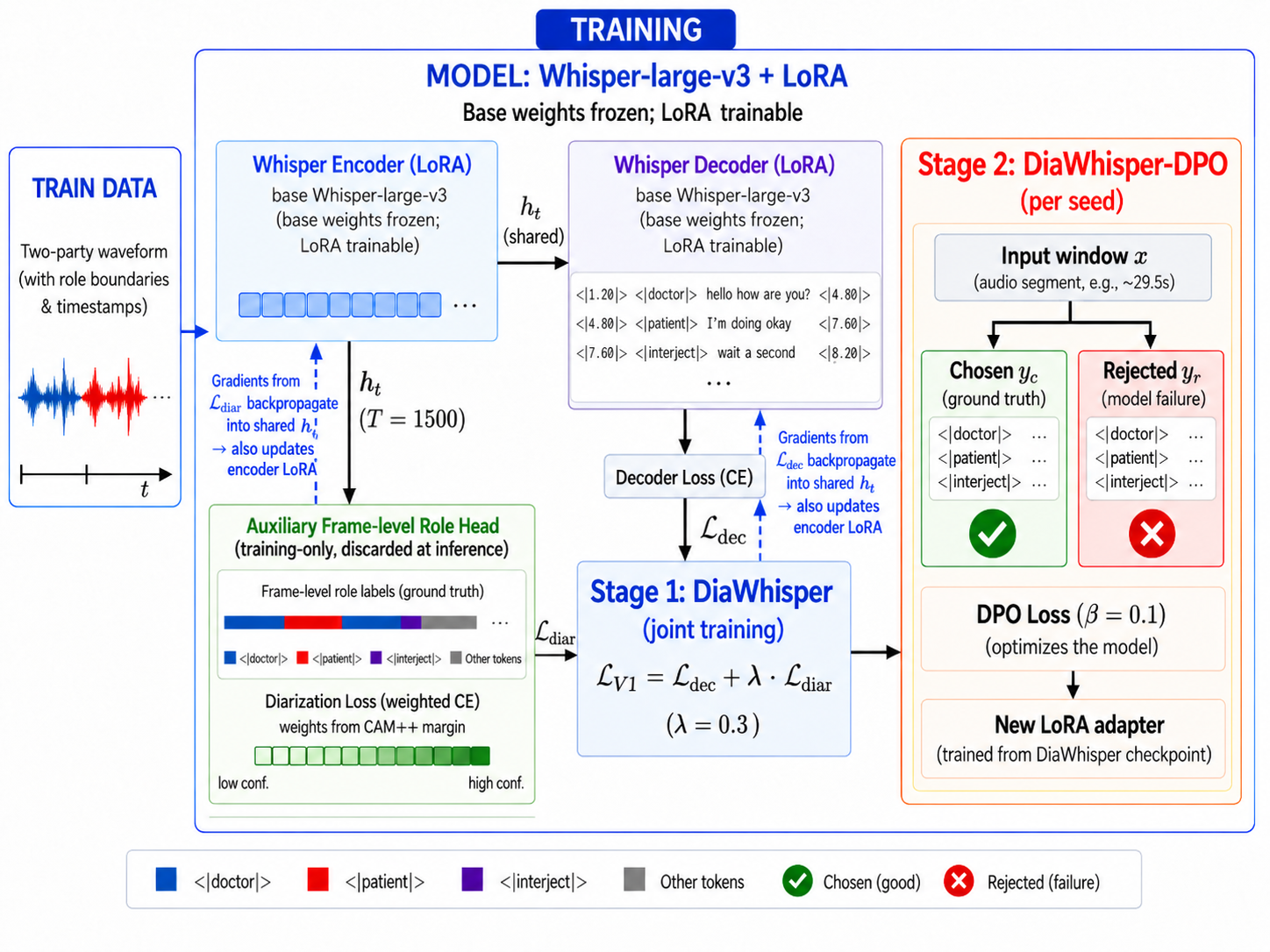}
\caption{Training pipeline: Stage~1 produces DiaWhisper; Stage~2 refines it into DiaWhisper-DPO. Both share the inference procedure in Section~\ref{ssec:inference}.}
\label{fig:ozwhisper_framework}
\end{figure*}

\vspace{-0.25cm}

\section{Method}
\vspace{-0.25cm}
We construct two-party clinical interview data, adapt Whisper for joint transcription, timestamps, and roles with an auxiliary diarization loss, then refine DiaWhisper using failure-mined DPO from training-set failures. Both models share the same inference procedure.
\label{sec:method}
\vspace{-0.25cm}

\subsection{Dataset Construction}
\label{ssec:dataset}

We use DAIC-WOZ \cite{gratch2014daic,ringeval2019avec} and the Chinese PDCH corpus \cite{pdch2025} as complementary clinical-interview corpora. DAIC-WOZ distributes only participant-side audio, so we reconstruct two-party dialogues by synthesizing Doctor and Patient speech from the ground-truth transcripts using XTTS \cite{casanova2024xtts,casanova2022yourtts}, preserving exact role boundaries and timestamps. For each session, two VCTK voices per role \cite{veaux2017cstr} are cross-combined into four variants, of which one is deterministically selected using SHA256(session\_id, filename), ensuring approximately balanced voice assignments without cross-split leakage. For Chinese validation, we use PDCH recordings containing depression consultations and Hamilton Depression Rating Scale (HAMD) assessment; speech is voice-converted for privacy, and our patient-level partitioning prevents speaker identity from crossing data splits.

\subsection{Architecture}
\label{ssec:architecture}

Fig.~\ref{fig:ozwhisper_framework} outlines the pipeline. We fine-tune Whisper-large-v3 \cite{radford2023whisper} with LoRA \cite{hu2022lora} (rank~=~16, $\alpha=32$, dropout~=~0.05) applied to self-attention query/value projections across both encoder and decoder. Targets follow Whisper's native timestamp--tag--text--timestamp format, replacing text spans with role-tagged spans (\texttt{<|doctor|>}, \texttt{<|patient|>}, or \texttt{<|interject|>} for overlap); only the embedding/output rows for the 3 role tokens are unfrozen.

We attach an auxiliary role head $g_\phi$ -- a single linear layer, 4-way: Doctor/Patient/Interject/None, training-only and discarded at inference -- to each encoder frame $h_t\in\mathbb{R}^{1280}$ ($t=1,\dots,T{=}1500$ frames per 30~s chunk). For each frame, a confidence weight $w_t$ comes from the margin $m_t$ -- the gap between how strongly frame $t$'s CAM++ \cite{chen20243dspeaker} speaker embedding $e_t$ resembles the session's Doctor versus Patient voiceprint centroid:
\begin{equation}
\begin{aligned}
m_t &= \cos(e_t, c_{Doctor}) - \cos(e_t, c_{Patient})\\
w_t &= f(|m_t|)
\end{aligned}
\label{eq:margin}
\end{equation}
where $f(\cdot)$ maps larger margins to higher confidence in 5 bins ($|m_t|$ thresholds 0.05/0.15/0.3/0.5), from 0.4 ($\sim$39\% measured reliability) to 1.0 ($\sim$99.8\%). This weight feeds an independent loss:
\begin{equation}
\mathcal{L}_{diar} = \frac{1}{T}\sum_{t=1}^{T} w_t \cdot \mathrm{CE}\big(g_\phi(h_t),\, r_t\big)
\label{eq:diar_loss}
\end{equation}
where $\mathrm{CE}$ is cross-entropy and $r_t$ the ground-truth frame role; low-confidence frames are down-weighted to limit ambiguous-frame gradients.

\subsection{Training}
\label{ssec:training}

In Stage 1 (DiaWhisper), we jointly optimize:

\begin{equation}
\mathcal{L}_{V1} = \mathcal{L}_{dec} + \lambda \cdot \mathcal{L}_{diar}, \quad \lambda = 0.3
\label{eq:v1_loss}
\end{equation}
where $\mathcal{L}_{dec}$ is teacher-forcing cross-entropy over the serialized target and $\lambda$ weights the auxiliary diarization loss. Training uses learning rate $1\times10^{-4}$, effective batch size 16, for 3 epochs across 3 seeds.

In Stage 2, we refine DiaWhisper with DPO \cite{rafailov2023dpo} to obtain \textbf{DiaWhisper-DPO}. From training windows triggering the 8-token repetition criterion, we sample about 500 per seed to form preference pairs (500/500/500 EN; 499/500/499 ZH), with the model's own genuine failure as rejected and ground truth as chosen. Stage-1 parameters are merged and frozen, and only a newly initialized policy LoRA adapter is optimized. With policy $\pi_\theta$, frozen reference $\pi_{ref}$, input $x$, chosen $y_c$, and rejected $y_r$:

\begin{align}
\Delta_\theta &= \log\pi_\theta(y_c|x)-\log\pi_\theta(y_r|x)
\label{eq:delta_theta}\\
\Delta_{ref} &= \log\pi_{ref}(y_c|x)-\log\pi_{ref}(y_r|x)
\label{eq:delta_ref}
\end{align}

\begin{equation}
\mathcal{L}_{DPO} = -\log\sigma\big(\beta[\Delta_\theta - \Delta_{ref}]\big)
\label{eq:dpo_loss}
\end{equation}
where $\sigma(\cdot)$ is the sigmoid and $\beta$ controls preference sharpness. Training uses $\beta=0.1$, learning rate $5\times10^{-5}$, for 1 epoch, mined and trained independently per seed.

\vspace{-0.25cm}

\vspace{-0.25cm}

\subsection{Inference and Decoding}
\label{ssec:inference}

Audio is segmented into $\sim$29.5s chunks, boundaries snapped to the nearest low-energy pause within $\pm2$s, since no ground-truth boundaries exist at inference. Each chunk is decoded greedily (beam search often loses the fine-tuned role/timestamp structure), stopping on 8 consecutive repeated tokens -- also the DPO-mining criterion (Section~\ref{ssec:training}) -- or a token budget. A repetition cutoff before full coverage triggers a retry (up to 4/chunk), resuming from an estimated point with sampling (temperature~=~0.5, top-p~=~0.9) instead of greedy, since resuming from near-identical context under greedy decoding reproduces the loop. Output is parsed into (role, start, end, text) triplets and concatenated per session; DiaWhisper and DiaWhisper-DPO share this procedure, differing only in checkpoint.

\section{Experimental Results and Analysis}

We assess whether failure-mined preference optimization improves the accuracy and robustness of role-attributed transcription across DAIC-WOZ and PDCH-HAMD, with additional analysis of seed sensitivity and decoding failures.

\subsection{Experimental Setup}
\label{ssec:setup}

DAIC-WOZ uses 132/28/29 train/dev/test sessions; our PDCH-HAMD subset has 100 sessions (49.8~h) from 73 patients, split patient-wise 51/11/11, yielding 26 test sessions. We compare three cascaded Whisper-large-v3 baselines sharing the same first-speaker heuristic and aligner, differing only in diarization: ECAPA-TDNN \cite{desplanques2020ecapa}, CAM++ \cite{chen20243dspeaker}, and pyannote.audio \cite{bredin2023pyannote}. Metrics include turn-level role accuracy after temporal alignment (unmatched reference turns count as incorrect), WER/CER, and Attribution Error Rate (AER) over matched reference turns. With $T_r$ denoting reference turns of role $r$ and $D_r=\{t\in T_r:\text{pred\_role}(t)\neq\text{Missing}\}$:

\begin{equation}
\text{AER}_r=
\frac{|\{t\in D_r:\text{pred\_role}(t)\neq r\}|}{|D_r|}
\label{eq:aer}
\end{equation}

AER therefore measures attribution conditional on non-missing turns; coverage errors are captured by DER. We also report NIST-style Diarization Error Rate (DER) \cite{fiscus2006rt06}:

\begin{equation}
\text{DER}=
\frac{\text{MD}+\text{SC}+\text{FA}}{\text{TOTAL}},
\label{eq:der}
\end{equation}

where MD is missed reference duration, SC misattributed duration, FA non-overlapping predicted duration, and TOTAL total reference speech. With only Doctor and Patient roles, SC directly reflects role errors, so DER is interpreted only within this task. We report 2000-resample session-bootstrap 95\% CIs for all systems; for DiaWhisper and DiaWhisper-DPO, \cite{efron1994bootstrap}, pooling three seeds within each sampled session. To quantify the DAIC-WOZ synthetic gap, we transcribe matched real and XTTS patient audio: mean WER is 0.136 vs.\ 0.106 across 189 sessions and 0.107 vs.\ 0.099 on the test split. PDCH-HAMD provides independent validation on real conversational interviews after voice conversion. Results appear in Tables~\ref{tab:main-accuracy}--\ref{tab:ablations}, with cross-dataset and stability analyses in Section~\ref{ssec:seedablation}.

\subsection{Main Results}
\label{ssec:mainresults}

Table~\ref{tab:main-accuracy} reports both datasets. DiaWhisper-DPO leads all baselines in role accuracy, DER, and AER. On DAIC-WOZ, the large WER gain mainly reflects recovered coverage after eliminating repetition loops that exhaust retries (Section~\ref{ssec:seedablation}). Its DER (0.119) is roughly one quarter of the strongest baseline's 0.431 and DiaWhisper's 0.485, while FA falls from $\sim$30\% for baselines to $\sim$10\%. Noisier, with untranscribed filler words and voice-converted speech, PDCH-HAMD shows the same ranking with a smaller margin: DiaWhisper-DPO achieves 0.313 DER versus Pyannote's 0.574, with FA of 8--17\%, suggesting errors are dominated by missed or misattributed speech. Per-role AER remains lowest for DiaWhisper-DPO at 0.01/0.02 on DAIC-WOZ and 0.26/0.26 on PDCH-HAMD. Oracle role mapping changes baseline accuracy by only 0--7 points (per-system values omitted for space); DiaWhisper-DPO still leads the best oracle baseline by 21.5 points on DAIC-WOZ and 17.1 on PDCH-HAMD, indicating mainly diarization rather than role-assignment errors.

\begin{table}[H]
\centering
\caption{Test-set results, DAIC-WOZ (n=29) and PDCH-HAMD (n=26). CI = session-bootstrap 95\% CI; DiaWhisper CIs pool 3 seeds/session. $\sigma$ = seed SD; AER (D/P) = Doctor/Patient.}
\scriptsize
\renewcommand{\arraystretch}{0.85}
\setlength{\tabcolsep}{0.8pt}
\begin{tabular}{llcccc}
\toprule
Method & Dataset & Role Acc. (CI, $\sigma$) & DER (CI) & WER/CER & AER(D/P) \\
\midrule
ECAPA       & DAIC & 0.464 [.39,.54] & 0.792 [.65,.95] & 0.104          & 0.20/0.60 \\
ECAPA       & PDCH & 0.480 [.46,.50] & 0.713 [.63,.81] & 0.489          & 0.41/0.62 \\
CAM++       & DAIC & 0.758 [.71,.80] & 0.431 [.37,.51] & 0.099          & 0.17/0.08 \\
CAM++       & PDCH & 0.507 [.49,.52] & 0.594 [.51,.69] & 0.489          & 0.55/0.42 \\
Pyannote    & DAIC & 0.600 [.56,.65] & 0.559 [.49,.64] & 0.099          & 0.43/0.17 \\
Pyannote    & PDCH & 0.514 [.46,.57] & 0.574 [.49,.66] & 0.628          & 0.28/0.65 \\
DiaWhisper  & DAIC & 0.645 [.62,.67] $\sigma$.205 & 0.485 [.46,.51] & 0.455$\pm$.175 & 0.09/0.02 \\
DiaWhisper  & PDCH & 0.570 [.55,.59] $\sigma$.052 & 0.447 [.39,.50] & 0.599$\pm$.111 & 0.33/0.45 \\
DiaW-DPO    & DAIC & \textbf{0.973 [.97,.98]} $\sigma$.002 & \textbf{0.119 [.10,.14]} & \textbf{0.048$\pm$.005} & 0.01/0.02 \\
DiaW-DPO    & PDCH & \textbf{0.757 [.74,.78]} $\sigma$.005 & \textbf{0.313 [.25,.38]} & \textbf{0.505$\pm$.073} & 0.26/0.26 \\
\bottomrule
\end{tabular}
\label{tab:main-accuracy}
\end{table}

\noindent

\subsection{Seed Stability and Error Analysis}
\label{ssec:seedablation}

On DAIC-WOZ, DiaWhisper is highly sensitive to the random seed (role accuracy 0.41--0.79, $\sigma$=.205), whereas DPO cuts this variation to $\sigma$=.002 (Table~\ref{tab:main-accuracy}). The DAIC-WOZ ablations (Table~\ref{tab:ablations}) show that removing retry drops DiaWhisper from 0.645 to 0.280 but barely moves DiaWhisper-DPO (0.973 to 0.971), and replacing DPO with SFT reaches only 0.684, so the improvement comes from the preference signal rather than retry recovery or extra exposure to ground truth. Replacing genuine failures with synthetic corruptions (role-swapped, truncated, looped, or mismatched) still recovers 0.910 role accuracy on DAIC-WOZ, but with 18$\times$ the seed variability ($\sigma$=.037 vs.\ .002), so genuine failures give a more stable refinement signal.

\begin{table}[H]
\centering
\caption{Ablations, DAIC-WOZ and PDCH-HAMD, mean$\pm$SD over 3 seeds. SFT and artificial negatives replace the DPO stage; CER replaces WER for PDCH-HAMD rows.}
\scriptsize
\renewcommand{\arraystretch}{0.90}
\setlength{\tabcolsep}{1.8pt}
\begin{tabular}{llcc}
\toprule
Variant & Dataset & Role Acc. & WER/CER \\
\midrule
SFT (replaces DPO)         & DAIC & 0.684$\pm$.182 & 0.441$\pm$.151 \\
SFT (replaces DPO)         & PDCH & 0.568$\pm$.085 & 0.720$\pm$.141 \\
DPO, artificial negatives  & DAIC & 0.910$\pm$.037 & 0.279$\pm$.075 \\
DPO, artificial negatives  & PDCH & 0.682$\pm$.049 & 0.443$\pm$.044 \\
DiaWhisper, no retry        & DAIC & 0.280$\pm$.127 & 0.761$\pm$.121 \\
DiaWhisper, no retry        & PDCH & 0.509$\pm$.089 & 0.625$\pm$.111 \\
DiaWhisper-DPO, no retry    & DAIC & 0.971$\pm$.002 & 0.047$\pm$.004 \\
DiaWhisper-DPO, no retry    & PDCH & 0.756$\pm$.004 & 0.481$\pm$.058 \\
\bottomrule
\end{tabular}
\label{tab:ablations}
\end{table}

\begin{figure}[h]
\centering
\includegraphics[width=0.7\linewidth]{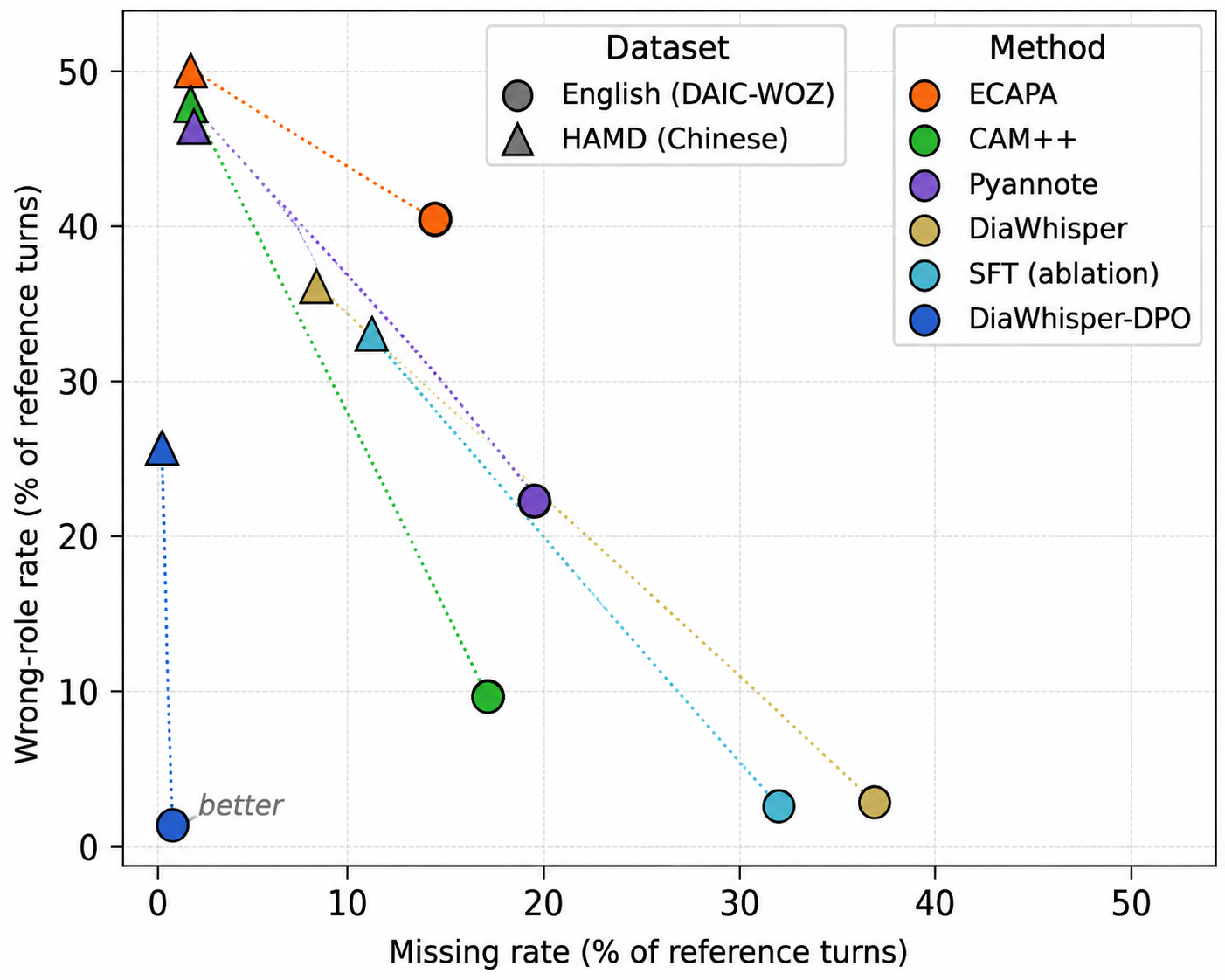}
\caption{Missing rate vs.\ wrong-role rate by method, DAIC-WOZ (circle) and PDCH-HAMD (triangle); dotted lines join each method's two points. DiaWhisper-DPO sits closest to the origin on both datasets, nearly eliminating missing turns on PDCH-HAMD as well as DAIC-WOZ, though wrong-role attribution remains the larger residual error there.}
\label{fig:error-analysis}
\end{figure}

Across all 87 matched DAIC-WOZ session-seed evaluations, DiaWhisper-DPO improves role accuracy by +0.220 to +0.442. On the full DAIC-WOZ test set (2694 chunks), retry exhaustion leaves 33.6\% of DiaWhisper's speech uncovered against 3.2\% for DiaWhisper-DPO, the main error source (Fig.~\ref{fig:error-analysis}); beyond this coverage gain, DAIC-WOZ matched-turn role accuracy also rises from 95.4\% to 98.3\% (non-overlapping 95\% CIs).

PDCH-HAMD follows the same pattern more weakly: seed variability falls from $\sigma$=.052 to .005, and the PDCH-HAMD ablations keep the ranking -- retry removal mildly hurts DiaWhisper (0.570 to 0.509) but not DiaWhisper-DPO (0.757 to 0.756), SFT (0.568) stays within DiaWhisper's seed noise, and artificial negatives (0.682) fall in between. DiaWhisper-DPO improves all 78 matched PDCH-HAMD session-seed evaluations (mean +0.188 role accuracy), with PDCH-HAMD matched-turn role accuracy rising from 62.3\% to 75.9\% (non-overlapping 95\% CIs); unlike DAIC-WOZ, the residual PDCH-HAMD error is concentrated in role attribution rather than coverage (0.2\% missing vs.\ 25.8\% wrong-role).

Finally, the real-versus-synthetic check covers only the DAIC-WOZ patient side, as no real interviewer audio is distributed, and since incorrect role attribution could distort depression-screening signals \cite{burdisso2024daicwoz}, such output still requires human verification.

\vspace{-0.35cm}

\section{Conclusion}
\label{sec:conclusion}

DiaWhisper-DPO fine-tunes Whisper-large-v3 with LoRA and an auxiliary role head, then applies DPO to genuine decoding failures for joint transcription and role attribution without separate diarization or human preference annotation. Independently trained on DAIC-WOZ and PDCH, it outperforms cascaded baselines and greatly reduces seed variability, demonstrating the effectiveness of failure-mined preference optimization for structured transcription.

\vspace{-0.35cm}

\section{Compliance with Ethical Standards}
PDCH was collected at Beijing Anding Hospital with ethics approval
(No.~2021-research-102) and written informed consent from all participants
\cite{pdch2025}. DAIC-WOZ was originally collected with USC IRB approval
(UP-11-00342) and informed consent; this study used the existing de-identified
dataset under its data-use terms \cite{gratch2014daic}.

{\footnotesize
\bibliographystyle{IEEEbib}
\bibliography{refs}
}

\end{document}